# ASSESSING THE FEASIBILITY OF LIGHTWEIGHT WHISPER MODELS FOR LOW-RESOURCE URDU TRANSCRIPTION

Abdul Rehman Antall*[1], Dr. Naveed Akhtar[2]
[1]National University of Computer and Emerging Sciences (FAST-NUCES), Lahore
[2]PTCL Group, Pakistan
L230605@lhr.nu.edu.pk
naveed.akhtar@ptclgroup.com

**ABSTRACT**

This study evaluates the feasibility of lightweight Whisper models (Tiny, Base, Small) for Urdu speech recognition in low-resource settings. Despite Urdu being the 10th most spoken language globally with over 230 million speakers, its representation in automatic speech recognition (ASR) systems remains limited due to dialectal diversity, code-switching, and sparse training data. We benchmark these models on a curated Urdu dataset using word error rate (WER), without fine-tuning. Results show Whisper-Small achieves the lowest error rates (33.68% WER), outperforming Tiny (67.08% WER) and Base (53.67% WER). Qualitative analysis reveals persistent challenges in phonetic accuracy and lexical coherence, particularly for complex utterances. While Whisper-Small demonstrates promise for deployable Urdu ASR, significant gaps remain. Our findings emphasize lay the groundwork for future research into effective, low-resource ASR systems.

***Index Terms***— Automatic speech recognition, Urdu, low-resource languages, pre-trained models, benchmarking

## 1. INTRODUCTION

Automatic Speech Recognition (ASR) is a technology designed to transform spoken language into written texts [1]. It has revolutionized human-computer interaction by enabling real-time transcription, voice-controlled systems, and accessibility tools [2]. Recent advancements in ASR, particularly self-supervised learning and transformer-based architectures, have significantly improved system efficiency and performance [3, 4]. Models like Whisper and Wav2Vec 2.0 have achieved near-human accuracy in many high-resource languages [5, 6].

However, low-resource languages such as Urdu continue to lag behind[7]. This inconsistency in performance is due to several factors. These include limited annotated data, phonetic complexity, and the lack of linguistic resources[8]. Furthermore, the majority of pre-trained ASR models are developed and optimized using large-scale datasets from high-resource languages such as English, Mandarin, or Spanish[9]. As a result, these models often struggle to generalize effectively to languages with different linguistic structures and sparse representation in training corpora.

Urdu ranks among the top ten most spoken languages in the world[10] with over 230 million speakers. Still, it remains underrepresented in both academic research and commercial ASR applications. The Urdu language is not a well-studied language [11] and its ASR training datasets are minimal [12]. This lack of representation raises important concerns about linguistic fairness, accessibility, and scalability of current ASR technologies. If ASR is to be a truly global tool, it must be evaluated and improved for languages that fall outside the scope of mainstream development[13].

Whisper is a multilingual ASR model developed by OpenAI. It was trained on 680,000 hours of multilingual and multitask supervised data [14] and is available in several model sizes which include Tiny, Base, Small, Medium, and Large. Even without fine-tuning, it functions effectively in a wide variety of languages. Because of this, it may be used with both high-end computers and low-power devices. Hence, It is an excellent choice for testing ASR in low-resource languages like Urdu because of its open-source nature and cross-lingual capabilities.

In this paper, the zero-shot capabilities of lightweight whisper models (whisper-tiny, whisper-base, and whisper-small) for the Urdu language are investigated. The results are intended to highlight these models' shortcomings, especially when there are limited computing resources and training data available.

## 2. RELATED WORK

### 2.1. ASR Robustness Across Languages and Conditions

Graham and Roll (2024) evaluated Whisper's ASR performance across native and non-native English accents, showing higher accuracy for American and Canadian speakers compared to British and Australian [15]. They also linked recognition errors to speaker traits such as L1 prosody, vowel inventory, and speech type (read vs conversational), highlighting Whisper's limitations with spontaneous and accented speech.

Katkov et al. (2024) conducted a multilingual study evaluating ASR performance under various audio challenges such as white noise, reverberation, time-stretching, and pitch shifts [16]. Their findings revealed a constant reduction in accuracy across English, Italian, and German, indicating that even cutting-edge models degrade dramatically in unfavorable settings. This underscores the need to evaluate ASR not just for language coverage, but also for resilience especially in low-resource or real-world deployment scenarios where audio quality cannot be guaranteed.

Huang et al. (2025) presented language embedding modeling based on linguistic similarities to increase Whisper's performance in unknown or underrepresented languages [17]. Their method shows that Whisper may be better tailored when linguistic connections are taken into account.

### 2.2. ASR for Low-Resource Languages

Many recent ASR advancements have focused on high-resource languages with rich datasets. However, for low-resource languages, there is a lack of significant labeled data. Yeroyan and Karpov (2024) described a unique method for generating ASR training datasets from audiobooks [18]. Their method aligns long audio files with text and segments them into shorter clips suitable for model training. While their work focuses on Armenian, their dataset creation technique is applicable to other low-resource languages and provides a solution for data scarcity.

### 2.3. Urdu-Specific ASR Research

The progression of Urdu ASR from traditional statistical techniques to modern pre-trained deep learning models was explored by Sharif et al. (2024) [12]. Their study underscores key challenges, including the scarcity of standardized datasets, orthographic complexity, and the limited inclusion of Urdu in multilingual pre-trained systems. Additionally, they examined current research trends, dataset development efforts, and architectural innovations, aiming to guide and inform future advancements in Urdu ASR.

Urdu is one of the world's most widely spoken languages, but it has minimal ASR research and tools. Arif et al. (2025) performed a large-scale benchmarking of Urdu ASR with Whisper, MMS, and SeamlessM4T[19]. According to their findings, Whisper outperformed SeamlessM4T in terms of conversational Urdu speaking. On the other hand, SeamlessM4T performed better on read speech. They also presented the first conversational Urdu ASR dataset and emphasized the need for greater text normalization in Urdu.

### 2.4. Gaps and Motivation

Most existing Urdu ASR research focuses on large-scale models like Whisper-Large and SeamlessM4T. Because these large models demand high-end GPUs and extensive memory, they are unsuitable for implementation in low-resource environments such as budget hardware or public-sector infrastructure in developing countries. Smaller Whisper models (Tiny, Base, and Small) are lighter and more cost-effective. However, their zero-shot performance in Urdu speech is little known.

This experiment seeks to assess these tiny models on genuine Urdu audio with no fine-tuning. The study is based exclusively on minimal computation and open-source tools, and it intends to guide developers looking for efficient, deployable ASR solutions for Urdu language situations.

## 3. METHODOLOGY

### 3.1. Overview

This study compares the performance of lightweight Whisper models to real-world Urdu speech data, with a focus on low-resource implementation practicality. The workflow relied entirely on open-source tools and publicly available models, allowing others to replicate the evaluation with ease.

### 3.2. Data Collection

Ten native Urdu speakers, four women and six men, were chosen from personal and social circles to provide a varied range of voices and speaking styles. Each participant was given a short set of Urdu sentences to read. These prompts were chosen to include a variety of sounds and common language patterns. Every speaker recorded four separate voice notes, resulting in 36 samples in total. Recordings were made in quiet indoor settings using personal smartphones and laptops. This setup reflects everyday conditions and devices typical in low-resource environments. All recordings were stored in .wav format with a sample rate of 16 kHz and the final speech dataset comprised approximately 10 minutes of recorded audio.

### 3.3. Model Selection and Technical Setup

This study made use of pre-trained voice recognition models, notably the Tiny, Base, and Small variations of OpenAI's Whisper architecture, which could be accessed through the Hugging Face Transformers library. The transcription pipeline was built in Python with a collection of well-known open-source modules. Tools and libraries used include:

- `transformers (Hugging Face)` – ran the pre-trained Whisper ASR models.
- `torch` – backend for model execution on CPU.
- `jiwer` – calculated WER.
- `ffmpeg-python` – handled audio format conversion.
- `FastAPI` – created the web interface.
- `Jinja2` – rendered HTML templates.

The experiments were conducted in an offline environment using Visual Studio Code. Moreover, no model fine-tuning was performed. The complete source code, audio corpus and processing pipeline are publicly available at: https://github.com/AbdulRehmanAntall/benchmarking-whisper-for-urdu.

### 3.4. System Configuration

All experiments were conducted on a personal machine equipped with an Intel Core i5 (8th Gen) processor, 8 GB RAM, and a 256 GB SSD, running Windows 10 with Python version 3.12.4. All three Whisper models ran smoothly on this setup without requiring specialized hardware.

### 3.5. Evaluation Metrics

Model performance was evaluated using two standard accuracy metrics: Word Error Rate (WER). These metrics quantify how closely the model's transcriptions match the original reference text. Prior to evaluation, all text was normalized to ensure consistency this included removing punctuation, and unifying spacing.The jiwer Python library was used to compute WER.

## 4. RESULTS AND DISCUSSION

### 4.1. Quantitative Statistics

**Table 1** summarizes the overall transcription performance of the Whisper models in terms of WER.While, Figures 1, 2, and 3 present line plots showing the variation of Word Error Rate (WER) across each individual voice sample used in the evaluation.

| Statistic | Tiny | Base | Small |
|---|---|---|---|
| Mean | 67.08 | 53.67 | 33.68 |
| Std Dev | 12.08 | 10.67 | 9.44 |
| CV (%) | 18.01 | 19.88 | 28.02 |
| Median | 67.12 | 53.57 | 32.12 |
| IQR | 10.59 | 11.08 | 9.87 |
| Min | 39.29 | 28.57 | 17.14 |
| Max | 95.79 | 79.31 | 57.50 |

**Table 1**: WER statistics for Whisper models

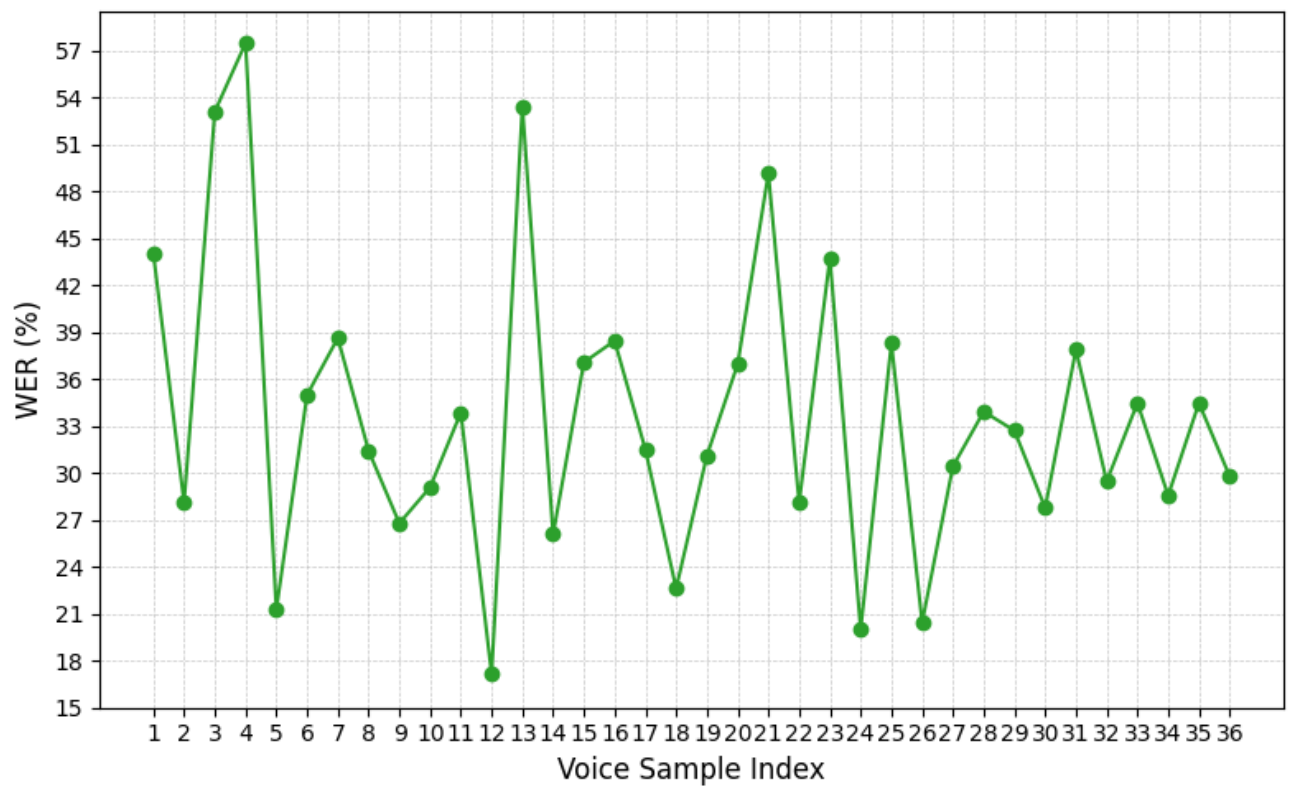

**Fig. 1**: WER per sample for Whisper-Small

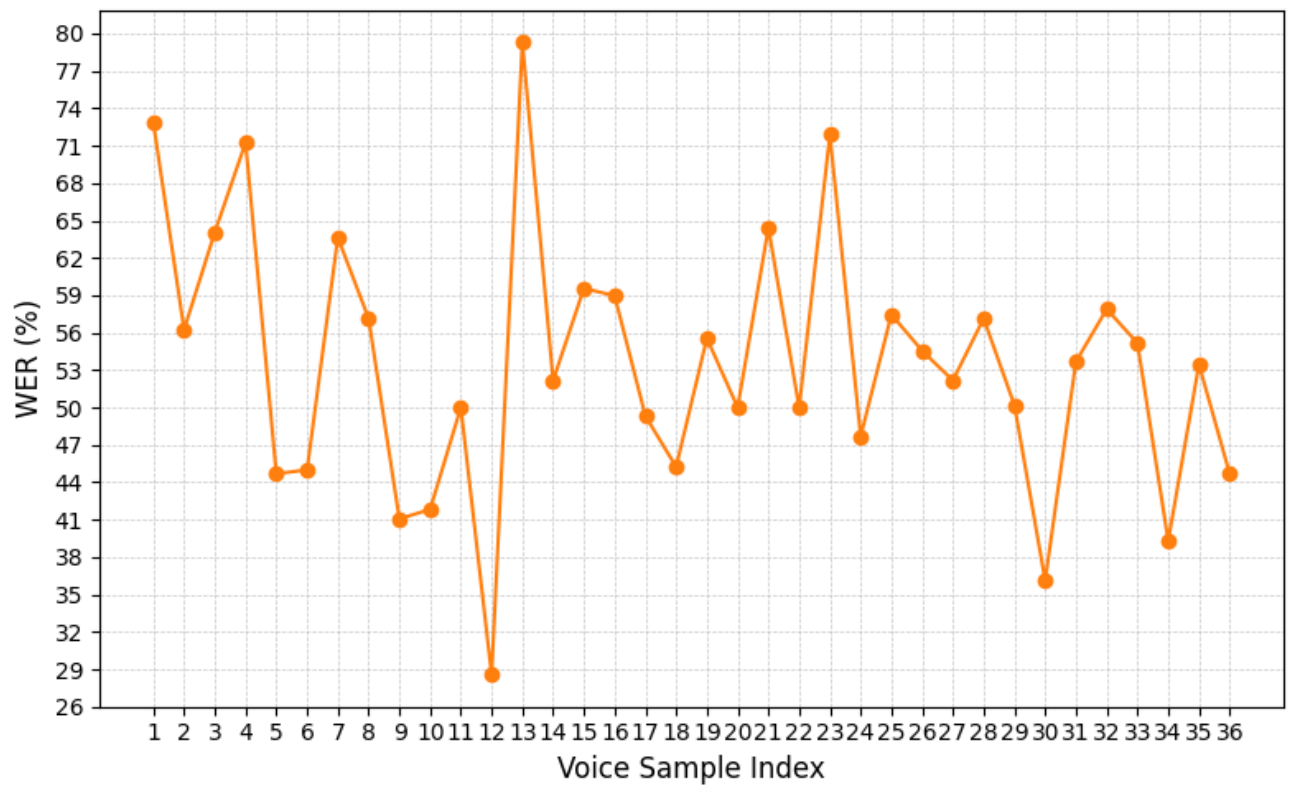

**Fig. 2**: WER per sample for Whisper-Base

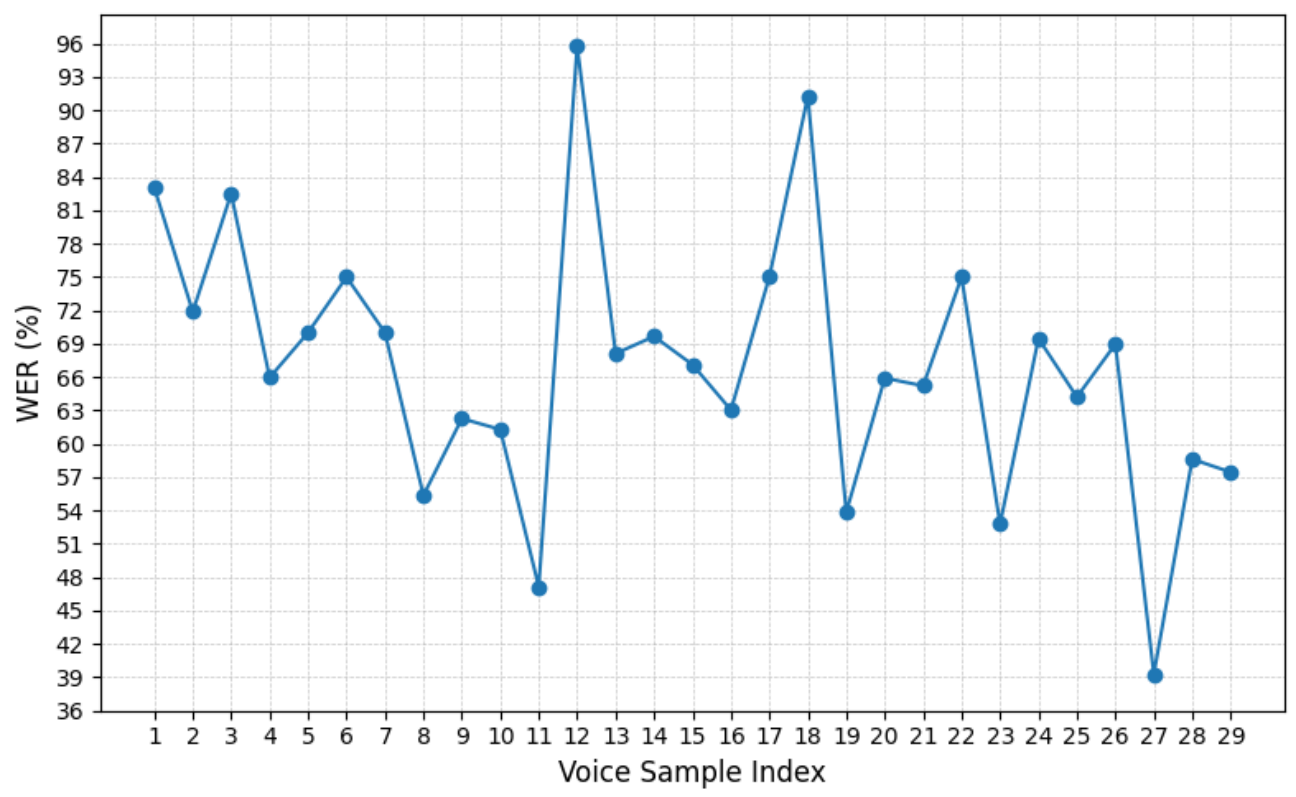

**Fig. 3**: WER per sample for Whisper-Tiny

Whisper-Small beats the other two models, with the lowest mean WER of 33.68%, indicating that it is more accurate in transcribing Urdu speech than the other two. Whisper-Base's performance is moderate, with WER value of 53.67%. Whisper-Tiny had the weakest performance, with a mean WER of 67.08%, indicating a limited capacity to handle the dataset. It is worth noting that none of the models achieved WERs of less than roughly 30%, which is considered a sufficient threshold for practical implementation.

Despite its relatively poor absolute performance, Whisper-Tiny has the lowest WER coefficient of variation, which is around 18%. This shows its consistently poor transcribing accuracy across samples. The low variability in Whisper-Tiny's performance suggests systematic errors that persist across different speaker styles, acoustic conditions, and utterance complexities. In contrast, Whisper-Small achieves higher average accuracy but exhibits greater performance fluctuation, with a coefficient of variation (CV) of approximately 28.02%. This increased variability implies that Whisper-Small's error rates are more sensitive to factors such as audio quality and speaker characteristics, despite its superior overall performance. Meanwhile, Whisper-Base achieves a modest reduction in errors compared to Whisper-Tiny, with a coefficient of variation (CV) of 19.88%, reflecting more stable but still non-negligible performance fluctuations.

Additionally, the close alignment between median and mean word error rate (WER) and character error rate (CER) across all models indicates symmetrical error distributions with minimal skewness.

### 4.2. Qualitative Analysis

To complement the quantitative metrics, a qualitative review was conducted across the these models. This analysis provides insights into transcription behaviour that are not fully captured by numerical scores. The focus is on recurring error patterns, model limitations, and architectural factors contributing to these issues. Example transcriptions and corresponding original prompts are provided in Annex A.

Across all models, several consistent challenges emerged. Common error types included:

- **Phonetic Substitutions:** Words with similar sounds were frequently confused, particularly among formal or Arabic-origin vocabulary.
- **Lexical Distortions:** Multi-syllabic or morphologically complex words were often misrepresented, especially in the Tiny and Base models.
- **Loss of Syntactic Integrity:** In longer sentences, the weaker models failed to preserve sentence structure or coherence.
- **Repetitive Artifacts:** In some cases, especially with Whisper-Tiny, the output included repeated syllables, suggesting issues in temporal attention or buffer overflow.

These errors are directly tied to the architectural and parameter constraints of the Whisper models. Whisper-Tiny (39M parameters) has limited representational capacity and shallow attention layers. This leads to frequent breakdowns in transcription. Whisper-Base (74M parameters) performs more consistently, but still struggles with morphological and contextual disambiguation. Whisper-Small (244M parameters), though not perfect, shows the strongest performance due to deeper layers and better phoneme recognition as reflected in its significantly lower WER.

The model's failure to generalize to formal Urdu or longer utterances suggests a mismatch between pretraining corpus distribution and the linguistic traits of Urdu. This aligns with quantitative trends as discussed in Section 4.1.

While Whisper-Small maintained the overall sentence structure in many cases, substitution errors (e.g., معاہدہ as موائدت) and misrecognitions of Arabic-rooted terms (e.g., بظاہر as بزائر) were still evident. By contrast, Whisper-Tiny often produced outputs that were either unintelligible or syntactically incoherent.

These observations reinforce the hypothesis that small ASR models are vulnerable to linguistic and structural complexity in low-resource languages like Urdu. The need for adaptation techniques or fine-tuning is apparent, especially when deploying such models in real-world environments.

Detailed transcription examples have been moved to Annex A for reference.

## 5. CONCLUSION

This paper presented a comprehensive evaluation of lightweight Whisper ASR models—Tiny, Base, and Small—for transcribing Urdu speech in a zero-shot setting. Quantitative analysis using Word Error Rate (WER) revealed that Whisper-Small delivered the best overall performance, achieving a mean WER of 33.68%. Whisper-Base showed moderate results, while Whisper-Tiny performed the weakest, with a mean WER of 67.08%.

Beyond the numbers, qualitative evaluation further confirmed these trends. Whisper-Small was able to preserve sentence meaning and structure even in long, lexically rich utterances, although it still produced minor phonetic and substitution errors. On the other hand, Whisper-Tiny often failed to produce coherent output, especially in complex or formal sentences, sometimes generating repetitive or unintelligible text.

In summary, Whisper-Small stands out as the most promising lightweight model for Urdu transcription in resource-constrained environments. However, even its performance indicates that there is still considerable room for improvement before such models can be reliably deployed in practical Urdu ASR applications.

## 6. FUTURE WORK

While this study focused on evaluating lightweight Whisper models in a zero-shot setting, future research could investigate the impact of fine-tuning these models on curated Urdu datasets to reduce WER and improve phonetic precision. Exploring adaptation techniques such as noise augmentation or domain-specific lexicons could further enhance robustness in real-world environments.

## A. SAMPLE TRANSCRIPTION EXAMPLES AND ERROR BREAKDOWN

This annex provides full transcription samples with detailed error annotations for each Whisper model. Error types are categorized as:

- **S:** Substitution – replacing a correct word with an incorrect one.
- **O:** Omission – skipping a word entirely.
- **R:** Repetition – repeating a word or syllable excessively.
- **D:** Distortion – heavily garbled or phonetically inconsistent transcription.

### Whisper-Small (Best Case Sample)

امریکہ اور یورپی یونین کے درمیان تجارتی معاہدہ طے پا گیا ہے جس کے تحت یورپی یونین سے امریکہ برآمد کی جانے والی تمام اشیا پر فیصد ٹیرف عائد ہو گا۔ اس معاہدے کے بعد بظاہر ایسا لگتا ہے کہ بیجنگ کے ساتھ ٹیرف پر بات چیت کے معاملے میں امریکہ کو برتری حاصل ہو سکتی ہے۔

امریکہ اور یورپی یونین کے درمیانت جارتی موائدتے پا گیا ہے جس کے تہت یورپی یونین سے امریکہ برامت کی جانے والی تمام احیوپر فیتت ترف آئت ہو گیاس موائدی کے بعد بزائر اسہ لکتا ہے کہ بیجن کے ساتھ ترف پر بات چیت کے مالے میں امریکہ کو بلتری حاصل ہو سکتی ہے۔

**Error Breakdown:**

| Actual Word | Transcribed As | Error Type |
|---|---|---|
| تجارتی | جارتی | S |
| معاہدہ | موائدت | S |
| برآمد | برامت | S |
| اشیا | احیوپر | D |
| فیصد | فیتت | S |
| ٹیرف | ترف | S |
| بظاہر | بزائر | S |
| ایسا | اسہ | S |
| بیجنگ | بیجن | S |
| معاملے | مالے | S |
| برتری | بلتری | S |

### Whisper-Base (Average Case Sample 2)

ڈیجیٹل دور میں مالیاتی لین دین تیز اور آسان بنانے کے لیے جاز کیش پاکستان میں سب سے زیادہ استعمال ہونے والی موبائل بینکنگ سروس بن چکی ہے۔ یہ سروس نہ صرف پیسے بھیجنے اور وصول کرنے میں مدد دیتی ہے بلکہ بل ادائیگی، آن لائن خریداری، قرض حاصل کرنے اور کاروباری لین دین میں بھی معاون ثابت ہوتی ہے۔ اگر آپ جاز کیش اکاؤنٹ بنانے، ایپ ڈاؤن لوڈ کرنے، جاز کیش بزنس اکاؤنٹ کھولنے یا پیسے بھیجنے کے طریقہ کے بارے میں جاننا چاہتے ہیں تو ہم اس مضموں میں تمام تفصیل دیں گے۔

دیجدل دورہ مالی آتی لینڈن تیڑ اور سان بنانے کے لئے جیاس کش پاکستان میں سب سے زیادہ استعمال ہونے والی مبائل بائل بینکنی سریس بن چکی ہے۔ یہ سریس نہ سے فسپے بھیجنے اور وہ سول کرنے میں مدتے تیئے بلکہ بلکہ بھیل آتاگی آنلین کڑے داری کرز حاصل کرنے اور کارو باری لینڈین نے بھی مابن سابت ہوتی ہے۔ اگر آپ جیاس کش ایکانٹ بنانے ایکانٹ کرنے جیاس کش بیسنس ایکانٹ کھولنے یا پیسے بھیجنے کے طریقے گے بارے میں جانا چاہتے ہیں تو ہم اس مزمون میں تمام تفسیل دیں گے۔

**Error Breakdown:**

| Actual Word | Transcribed As | Error Type |
|---|---|---|
| ڈیجیٹل | دیجدل | S |
| مالیاتی | مالی آتی | S |
| لین دین | لینڈن | S |
| تیز | تیڑ | S |
| آسان | سان | D |
| جاز کیش | جیاس کش | S |
| موبائل | مبائل | S |
| بینکنگ | بینکنی | S |
| سروس | سریس | S |
| پیسے | فسپے | S |
| وصول | وہ سول | S |
| مدد دیتی ہے | مدتے تیئے | D |
| بل ادائیگی | بھیل آتاگی | D |
| آن لائن | آنلین | S |
| خریداری | کڑے داری | S |
| قرض | کرز | S |
| کاروباری | کارو باری | S |
| لین دین | لینڈین | S |
| معاون | مابن | S |
| ثابت | سابت | S |
| اکاؤنٹ | ایکانٹ | S |
| ایپ ڈاؤن لوڈ | ایکانٹ کرنے | D |
| بزنس اکاؤنٹ | بیسنس ایکانٹ | S |
| طریقہ | طریقے | S |
| کے بارے میں | کے گے بارے میں | I |
| مضموں | مزمون | S |
| تفصیل | تفسیل | S |

## Whisper-Tiny (Moderate Failure Case)

نیوز رپورٹ میں کہا گیا ہے کہ منشیات کی سمگلنگ، انتہا پسندانہ خیالات کی حوصلہ افزائی، سیاسی پناہ کے حصول کے لیے غلط دستاویزات پیش کرنے، امیگریشن کے ضوابط کی خلاف ورزی، تیسرے ملک کی شہریت رکھنے اور تاجکستان کی سرزمین کو ٹرانزٹ ملک کے طور پر استعمال کرنے جیسے جرائم میں کچھ غیر ملکی شہریوں کے ملوث ہونے کی تصدیق کی گئی ہے۔

لیوی ربات میں کہا ہگی ہے کہ من چیاد کیس مگلیں دہا پر سندانہ کے علاتی ہو سلا آفزائی سیاسی پاناکتی حسول کے لیکوالہ دستہ ویداد پیشکہ نہیں مگرشن کے زواباتی باپرزی اسم ملکی شاری انتر اپنے ملکی ستان کی سنتمیل ک رانسل ملک کے دور پر استمال کرنے جسے جرائے میں کوشکر ملکی شاریوں پر ملہ مصانکی تستیقی ہے۔

**Error Breakdown:**

| Actual Word | Transcribed As | Error Type |
|---|---|---|
| منشیات | من چیاد | S |
| سمگلنگ | س مگلیں | D |
| انتہا پسندانہ | سندانہ | D |
| خیالات | علاتی | S |
| حصول | حسول | S |
| دستاویزات | دستہ ویداد | D |
| امیگریشن | مگرشن | D |
| ضوابط | زواباتی | S |
| استعمال | استمال | S |
| ملک | ملکی ستان | R |

## Whisper-Tiny (Complete Failure Case)

ملک کی معیشت پر پڑنے والے اضافی بوجھ کو بھی کم کیا جا سکے گا تاہم عام صارفین پر اس کا فوری اثر دیکھنے میں آئے گا کیونکہ انہیں مہنگی بجلی کا سامنا کرنا پڑے گا۔

ایک ایک ایک ایک ایک ایک ایک ایک ایک ایک ایک ایک ایک ایک ایک ایک ایک ایک ایک ایک ایک ایک ایک ایک ایک ایک ایک ایک ایک ایک ایک ایک ایک ایک ایک ایک۔

**Error Breakdown:**

| Actual Word | Transcribed | Error Type |
|---|---|---|
| *All content* | ایک ایک ... | R (Looped repetition) |
| Entire sentence | – | O |
| Context | – | D |